\documentclass[runningheads]{llncs}
\usepackage[T1]{fontenc}
\usepackage{graphicx,verbatim, amssymb, tabularx, xcolor, colortbl, booktabs}
\usepackage{tikz, overpic, amsmath}
\usetikzlibrary{spy, positioning, calc, math}
\usepackage[para]{footmisc}
\newcolumntype{C}{>{\centering\arraybackslash}X}

\begin{document}
\title{Test-Time Adaptation in Optical Coherence Tomography Using Trajectory-Aligned Time-Independent Flow}
\titlerunning{TTA-Flow}
%
\author{Veit Hucke\thanks{Equal contribution.}\inst{1,2}\orcidID{0009-0002-1519-1252} \and
Thomas Pinetz\protect\footnotemark[1]\protect\footnote[2]{Corresponding Author}\inst{1,2}\orcidID{0000-0002-6100-2136} \and
Gregor Reiter\inst{3}\orcidID{0000-0001-7661-4015} \and
Ursula Schmidt-Erfurth\inst{4}\orcidID{0000-0002-7788-7311} \and
Hrvoje Bogunovi{\'c}\inst{1,2}\orcidID{0000-0002-9168-0894}
}

\authorrunning{Hucke et al.}
%
\institute{Institute of Artificial Intelligence, Center for Medical Data Science, Medical University of Vienna, Austria \\
\and
Comprehensive Center for Artificial Intelligence in Medicine, Medical University of Vienna, Austria \\
\and
Department of Ophthalmology and Optometry, Medical University of Vienna, Austria\\
\and
Laboratory for Ophthalmic Image Analysis, Medical University of Vienna, Austria\\\email{thomas.pinetz@meduniwien.ac.at, hrvoje.bogunovic@meduniwien.ac.at}
}

  
\maketitle              
\begin{abstract}
Optical coherence tomography (OCT) is essential in ophthalmology, but inconsistent image quality especially in low-cost devices hinders automated analysis. To address this, we introduce a flow-matching-based test-time adaptation method that generates high-quality surrogate images from noisy inputs. Typically, domain gaps between test and training data cause pixel distribution mismatches during the denoising process. We overcome this by matching the test image's histogram to synthetic reference trajectories, successfully aligning the input with expected distributions. Additionally, we remove the network's time conditioning to account for slight deviations in real-world noise distributions. Our approach achieves state-of-the-art performance in segmenting critical biomarkers for two stages of Age-related Macular Degeneration (AMD). Code is available here: \url{https://github.com/Veit21/tta-flow}.


\end{abstract}

\section{Introduction}

Deep learning-based medical image segmentation has become a clinical staple for tracking biomarkers and detecting lesions~\cite{AzAg24}. However, model performance often deteriorates significantly when encountering minor distribution shifts, such as changes in imaging hardware~\cite{GoKi25}. This vulnerability is especially problematic with low-cost scanners, which introduce more severe noise characteristics and are rarely represented in standard training datasets.

To address this, researchers have increasingly turned to test-time adaptation (TTA)~\cite{FaPi25,SaNi25,LiDu24}. Early methods updated pre-trained model weights during inference using target domain data and specialized loss functions~\cite{wang2021tentfullytesttimeadaptation,wang2022continualtesttimedomainadaptation,PiHu26}. These approaches typically relied on prediction entropy (e.g., TENT~\cite{wang2021tentfullytesttimeadaptation}) or generating pseudo-labels via augmentations and thresholding (e.g., COTTA~\cite{wang2022continualtesttimedomainadaptation}). However, developing loss functions to provide consistent improvement across diverse problems without access to the training distribution is difficult, and recent findings indicate that their performance improvements are often inconsistent, especially in segmentation tasks~\cite{YiCh24}.

As an alternative, generative models have been used to adapt the test images directly, effectively projecting them onto the training data manifold for downstream tasks~\cite{LiLi24}. While earlier attempts utilized GANs~\cite{HoTz18} and normalizing flows~\cite{SaNi25}, diffusion models are now predominantly used for this task~\cite{LiDu24,gao2023sourcediffusiondriventesttimeadaptation,sanqian2023contentpreservingdiffusionmodelunsupervised,li2024score}. These methods generally treat the incoming test image as an intermediate point on a diffusion trajectory and reverse it, often incorporating a data-fidelity term to enforce consistency~\cite{sanqian2023contentpreservingdiffusionmodelunsupervised,LiDu24,FaPi25}. The primary flaw in this approach is that real-world noise is rarely ideal, fixed-magnitude Gaussian noise. Methods like DDA attempt to compensate by re-noising images to better fit the trajectory and applying data fidelity on downsampled versions~\cite{gao2023sourcediffusiondriventesttimeadaptation}. Unfortunately, downsampling risks blurring or eliminating small lesions, critical for diagnosis.
Others have incorporated large language models and CLIP-based method to synthesize realistic surrogates~\cite{LiLi24}, however this requires disease specific models that understand the modality well.

Recently, flow matching~\cite{lipman2023flowmatchinggenerativemodeling,lipman2024flowmatchingguidecode} has emerged as a robust alternative to classical diffusion models like DDPM~\cite{ho2020ddpm} improving the state-of-the in natural images~\cite{MaGa25} and in the medical domain~\cite{YaMe25}.
While variational setups have successfully combined flow matching with data-consistency priors for natural image restoration~\cite{MaGa25}, they still assume purely synthetic degradataions.
To make flow matching viable for real-world low-cost devices, we introduce two key modifications:
\begin{enumerate}
    \item \textbf{Trajectory Alignment:} We sample a reference trajectory histogram and use histogram matching to align the incoming test image, effectively bridging the gap between theoretical and actual noise distributions.
    \item \textbf{Time-Independent Flow:} Because actual noise levels deviate from the reference trajectory, we intentionally omit time conditioning in the flow network. As demonstrated by He et al.~\cite{sun2025noiseconditioningnecessarydenoising}, allowing the network to implicitly manage unknown real-world noise levels, rather than forcing a specific time condition, yields superior results.
\end{enumerate}

We demonstrate the effectiveness of our approach, TTA-Flow, on two clinical biomarkers for Age-related Macular Degeneration (AMD) in both 3D-to-3D and 3D-to-2D segmentation tasks using two different low-cost Optical Coherence Tomography (OCT) devices.

\section{Methods}

\begin{figure}[tb]
    \includegraphics[width=\textwidth]{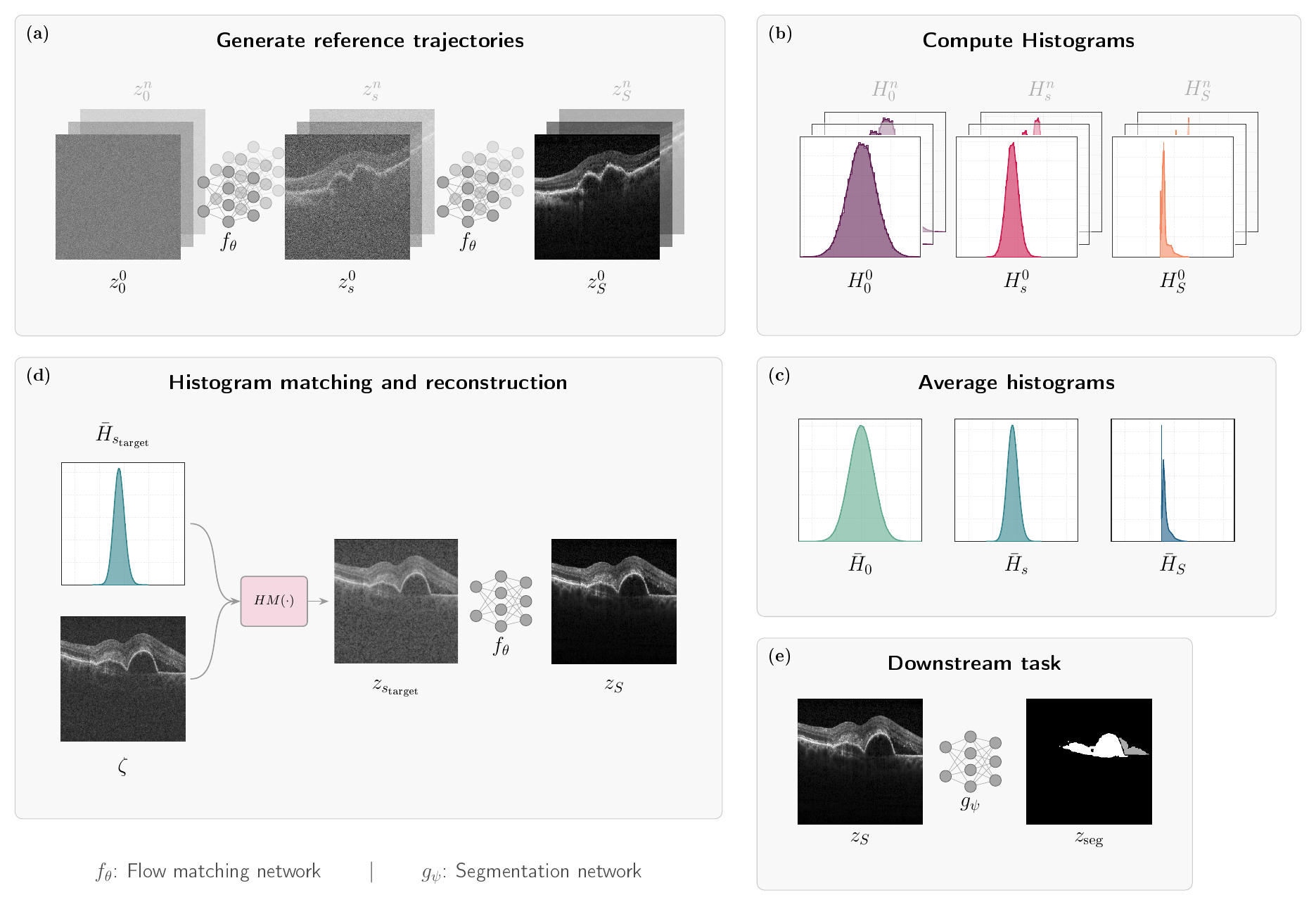}
    \caption{Schematic overview of the TTA-Flow framework. The process begins by (a) generating reference trajectories and (b, c) calculating an average histogram $\bar{H}_s$ for each time step. During inference, (d) an incoming test sample $\zeta$ is matched to the corresponding reference histogram before being processed by the flow matching network. Finally, (e) the reconstructed sample $z_S$ is utilized for downstream evaluation.}
    \label{fig:visual_abstract}
\end{figure}

A comprehensive visual overview of our proposed approach, TTA-Flow, is provided in Fig.~\ref{fig:visual_abstract}.

\subsubsection{Flow Matching for Image Reconstruction}

Building on the generative flow-matching (FM) framework introduced by Lipman et al.~\cite{lipman2023flowmatchinggenerativemodeling,lipman2024flowmatchingguidecode}, our approach learns a time-dependent vector field $\mathbf{v}_\theta$. This vector field is designed to transport a known Gaussian source distribution $p_0\sim \mathcal{N}(\mathbf{0}, \mathbf{I})$ to the empirical data distribution of our training set, $y_i \in \mathcal{D}$ for $i \in [0, N]$.
This transport is defined by a deterministic flow $\{\mathbf{z}_t\}_{t\in[0,1]}$ governed by the ordinary differential equation (ODE):
\begin{equation}\label{eqn:VectorField}
    \frac{d\mathbf{z}_t}{dt}=\mathbf{v}_t(\mathbf{z}_t), \qquad \mathbf{z}_0\sim p_0,
\end{equation}
which induces a probability path $p_t$ satisfying the continuity equation $\partial_t p_t(\mathbf{z}) + \nabla\cdot(p_t(\mathbf{z})\mathbf{v}_t(\mathbf{z}))=0$. 

To create a tractable training objective, flow matching requires defining a specific probability path between the source noise $\mathbf{z}_0$ and a target data sample $\mathbf{y}_i$.
We utilize a deterministic linear interpolation for this path:
\begin{equation}\label{eqn:LinearPath}
    \mathbf{z}_t = (1-t)\mathbf{z}_0 + t\mathbf{y}_i, \qquad t\in[0,1],
\end{equation}

This formulation establishes the target velocity as $\mathbf{u}_t(\mathbf{z}_t\,;\mathbf{z}_0,\mathbf{y}_i) = \mathbf{y}_i - \mathbf{z}_0$.
The network is then trained to minimize the regression loss against this target vector field:
\begin{equation}\label{eqn:FlowMatchingObjectiveFunction}
    \mathcal{L}_{\mathrm{FM}}=\mathbb{E}_{t,\mathbf{x}_0,\mathbf{x}_1}\left[\left\|\mathbf{v}_\theta(\mathbf{z}_t)-\mathbf{u}_t(\mathbf{z}_t\,;\mathbf{z}_0,\mathbf{y}_i)\right\|_2^2\right].
\end{equation}

Rather than predicting the velocity directly, we parameterize our model to predict the target sample $\mathbf{x}_{\theta}(\mathbf{z}_t) \approx \mathbf{y}_i$~\cite{li2026basicsletdenoisinggenerative}. The corresponding velocity can then be recovered via:
\begin{equation}\label{eqn:XPredToVPred}
    \mathbf{v}_{\theta}(\mathbf{z}_t)=\frac{\mathbf{x}_{\theta}(\mathbf{z}_t)-\mathbf{z}_t}{1-t}.
\end{equation}

This recovered velocity $\mathbf{v}_{\theta}$ is then substituted back into the objective function $\mathcal{L}_{\mathrm{FM}}$.
At test time, we conceptualize the noisy input sample $\zeta$ as an intermediate point interpolated between pure noise and a theoretical high-quality acquisition $y_k$ (e.g., $\zeta \approx tz_k + (1-t) y_k$). However, the true noise characteristics of $\zeta$ rarely match the exact theoretical noise of a random trajectory at time $t$.
Because of this discrepancy, we intentionally omit explicit noise-level (or time) conditioning from the network.
As demonstrated by He et al.~\cite{sun2025noiseconditioningnecessarydenoising}, removing this constraint prevents the network from forcing inaccurate assumptions and yields better results when the true noise level of the test image is unknown.

During inference, we generate novel samples by solving the initial value problem starting from pure noise with an Euler solver for $S\in \mathbb{N}$ steps:
\begin{equation}\label{eqn:InferenceODE}
    \mathbf{z}_{i + 1} = \mathbf{z}_i +  \frac 1 S \mathbf{v}_\theta(\mathbf{z}_i), \qquad \mathbf{z}_0 \sim \mathcal{N}(\mathbf{0},\mathbf{I})
\end{equation}

\subsubsection{Reference generation}

As previously noted, the underlying statistical properties of a test sample $\zeta$ will inherently differ from the trajectories observed during training.
To account for this discrepancy, we use our trained flow matching model to simulate $n$ sample trajectories according to Eq.~\ref{eqn:InferenceODE}.
From the resulting $Sn$ generated images, we calculate the mean target intensity histograms, denoted as $\{\bar{H}_s\}_{s\in[1,S]}$, for each discrete time step.
This sequence of average histograms establishes the reference trajectory for our specific flow model (Fig.~\ref{fig:visual_abstract}b, c).

\subsubsection{Inference}

During inference, we utilize this pre-computed sequence of average target histograms $\{\bar{H}_s\}_{s\in[1,S]}$ to normalize the incoming test data.
For a given test sample $\zeta$, we apply histogram matching as a preprocessing step.
This aligns the sample's intensity distribution with a specific reference histogram $\bar{H}_{s_{\text{target}}}$, where $s_{\text{target}}$ is a predetermined hyperparameter.
This alignment produces an intermediate sample that closely mirrors the expected data distribution of the reference trajectories at time $s_{\text{target}}$ (Fig.~\ref{fig:visual_abstract}d).
To obtain the clean reconstructed estimate $z_S$, we initialize the ODE (Eq.~\ref{eqn:InferenceODE}) using $z_s$ at step $s = s_{\text{target}}$.
We then execute the remaining $S - s_{\text{target}}$ iterations of the flow matching process (Eq.~\ref{eqn:InferenceODE}), effectively mapping the image into the target distribution.
Notably, our experiments demonstrate that this approach performs robustly even without the inclusion of explicit data-fidelity terms.
Finally, the reconstructed image $z_S$ is passed to downstream tasks to produce the final output $z_\text{seg}$ (Fig.~\ref{fig:visual_abstract}e).

\section{Experimental Setup}
\textit{Datasets:} We evaluate our approach on two Optical Coherence Tomography (OCT) datasets containing distinct clinically relevant pathologies and multiple scanners.
First, the public \textit{RETOUCH}~\cite{bogunovic2019retouch} dataset contains annotations for intraretinal fluid (IRF), subretinal fluid (SRF), and pigment epithelial detachment (PED). It comprises 70 total volumes acquired across three devices: Spectralis (24 vols, 49 B-scans/vol), Topcon (22 vols, 128 B-scans/vol), and Cirrus (24 vols, 128 B-scans/vol). 
Second, we utilize an \textit{in-house paired dataset}~\cite{Eidenberger2026multidevice} of 40 patients collected at Medical University of Vienna diagnosed with geographic atrophy (GA), a late stage of Age-related macular degeneration (AMD). For each patient, paired volumes were acquired using both Topcon (128 B-scans/vol) and Spectralis (49 or 97 B-scans/vol) scanners, with  manual annotations of GA provided on every alternating B-scan. We consider Topcon and Cirrus to be low-cost devices and adapt their images towards the Spectralis device, which has notably higher SNR.
To establish a controlled baseline and ensure reproducibility, all generative models are trained exclusively on the RETOUCH Spectralis subset. We restrict the quantitative analysis to the annotated B-scans.

\textit{Flow Matching Setup:} We employ a U-Net backbone~\cite{ronneberger2015unetconvolutionalnetworksbiomedical} ($\sim$103M parameters) for the flow-matching network. Models are trained for $300{,}000$ iterations using the Adam optimizer (learning rate $10^{-4}$, weight decay $10^{-2}$, batch size 16) with an exponential moving average (EMA) decay of 0.999. All input B-scans are resized to $224 \times 224$ pixels, normalized to $[-1,1]$, and augmented with random horizontal flipping.
During inference, the probability-flow ODE (Eq.~\ref{eqn:InferenceODE}) is integrated over $S=100$ steps.

\textit{Downstream Segmentation Models:} For fluid lesion segmentation, we employ a UNet++ architecture~\cite{zhou2018unetnestedunetarchitecture++} featuring a ResNet-18~\cite{he2015deepresiduallearningimage} encoder. To accurately gauge the quality of the reconstruction algorithms, this downstream model is trained solely on the RETOUCH Spectralis subset. We establish device-specific upper performance bounds (Upper) for the Topcon and Cirrus scanners by training dedicated, supervised models and evaluating them via 4-fold cross-validation. For GA segmentation, we use a pre-trained 3D-to-2D CNN based on~\cite{morano2023ssl}. Because our in-house dataset contains paired scans, we can directly compute the Spectralis upper bound without requiring further training.

\textit{Baselines and Hardware:} We compare our approach against classical test-time adaptation strategies (TENT~\cite{wang2021tentfullytesttimeadaptation} and CoTTA~\cite{wang2022continualtesttimedomainadaptation}) and two diffusion-based models (CPDM~\cite{sanqian2023contentpreservingdiffusionmodelunsupervised} and DDA~\cite{gao2023sourcediffusiondriventesttimeadaptation}). Additionally, we include a baseline using standard flow matching~\cite{li2026basicsletdenoisinggenerative} without our modifications. Hyperparameters for all methods were chosen by grid search; for our method, this involved selecting the optimal target time point $s_{\text{target}}$. All models were implemented in PyTorch and trained on a single NVIDIA RTX A6000 (48GB) GPU. Training the flow-matching model took approximately 80 hours. Inference time per B-scan averages 2.24 and 2.20 seconds with and without histogram matching. Generating the initial $n=100$ reference trajectories is a one-time process that takes roughly 231 seconds.

\begin{table}[t!]
\centering
\caption{Downstream lesion segmentation performance on Cirrus $\to$ Spectralis and Topcon $\to$ Spectralis. Dice similarity coefficient (DSC; higher is better) for IRF, SRF, PED, and GA. FID (lower is better) is also reported for Topcon.* denotes statistical significance (wilcoxon test $p<0.05$) of Ours\textsubscript{uncond.} and its closest competitor.}\label{tab:merged_performance}
\begin{tabularx}{\textwidth}{X C C C C C C}
\hline
 & \multicolumn{4}{c}{\textbf{RETOUCH}} & \multicolumn{2}{c}{\textbf{In-house}} \\
\cmidrule(lr){2-5}\cmidrule(lr){6-7}
\textbf{Method} & \textbf{IRF}$\uparrow$ & \textbf{SRF}$\uparrow$ & \textbf{PED}$\uparrow$ & \textbf{Mean}$\uparrow$ & \textbf{GA DSC}$\uparrow$ & \textbf{FID}$\downarrow$\\
\hline
\multicolumn{7}{c}{\textbf{Cirrus $\to$ Spectralis}} \\
\hline
\rowcolor{gray!10} Baseline & $51.7 \pm 27.0$ & $43.2 \pm 31.0$ & $22.5 \pm 32.1$ & $39.1$ & -- & -- \\
CPDM~\cite{sanqian2023contentpreservingdiffusionmodelunsupervised} & $55.2 \pm 27.7$ & $48.5 \pm 29.9$ & $28.9 \pm 33.2$ & $44.2$ & -- & -- \\
\rowcolor{gray!10} DDA~\cite{gao2023sourcediffusiondriventesttimeadaptation} & $50.0 \pm 28.3$ & $45.2 \pm 30.5$ & $25.6 \pm 32.5$ & $40.3$ & -- & -- \\
FM~\cite{li2026basicsletdenoisinggenerative} & $57.1 \pm 24.6$ & $56.3 \pm 24.6$ & $39.1 \pm 30.7$ & $50.8$ & -- & -- \\
\rowcolor{gray!10} CoTTA~\cite{wang2022continualtesttimedomainadaptation} & $52.0 \pm 22.3$ & $64.5 \pm 19.0$ & $36.5 \pm 28.6$ & $51.0$ & -- & -- \\
TENT~\cite{wang2021tentfullytesttimeadaptation} & $51.5 \pm 22.2$ & $64.4 \pm 19.0$ & $35.5 \pm 27.6$ & $50.5$ & -- & -- \\
\hline
\rowcolor{gray!10} Ours\textsubscript{uncond.} & $\mathbf{58.8} \pm 18.6$ & $\mathbf{66.3} \pm 17.2$ & $\mathbf{50.6} \pm 27.0$ & $\mathbf{58.6}^\ast$ & -- & -- \\
Ours\textsubscript{cond.} & $\underline{57.4} \pm 18.5$ & $\underline{64.7} \pm 17.8$ & $\underline{45.8} \pm 27.9$ & $\underline{56.0}$ & -- & -- \\
\hline
\rowcolor{gray!10} Upper & $69.5 \pm 17.0$ & $63.1 \pm 27.6$ & $52.1 \pm 35.3$ & $61.6$ & -- & -- \\ 
\hline
\multicolumn{7}{c}{\textbf{Topcon $\to$ Spectralis}} \\
\hline
\rowcolor{gray!10} Baseline & $49.9 \pm 26.0$ & $42.3 \pm 31.0$ & $28.0 \pm 18.8$ & $40.0$ & $43.7 \pm 24.8$ & $550.5$\\
CPDM~\cite{sanqian2023contentpreservingdiffusionmodelunsupervised} & $43.8 \pm 26.4$ & $45.3 \pm 31.0$ & $38.2 \pm 22.0$ & $43.8$ & $52.4 \pm 23.2$ & $100.1$\\
\rowcolor{gray!10} DDA~\cite{gao2023sourcediffusiondriventesttimeadaptation} & $38.0 \pm 27.7$ & $33.6 \pm 28.5$ & $30.6 \pm 18.0$ & $34.1$ & $53.2 \pm 21.7$ & $148.4$\\
FM~\cite{li2026basicsletdenoisinggenerative}  & $53.3 \pm 25.4$ & $49.3 \pm 25.9$ & $42.5 \pm 26.7$ & $48.3$ & $49.4 \pm 25.0$ & $131.2$\\
\rowcolor{gray!10} CoTTA~\cite{wang2022continualtesttimedomainadaptation} & $\underline{53.5} \pm 21.7$ & $47.7 \pm 23.4$ & $32.7 \pm 21.0$ & $44.6$ & N/A & N/A\\
TENT~\cite{wang2021tentfullytesttimeadaptation} & $\mathbf{53.9} \pm 22.4$ & $47.9 \pm 24.5$ & $33.3 \pm 20.4$ & $45.0$ & $48.6 \pm 31.9$ & N/A\\
\hline
\rowcolor{gray!10} Ours\textsubscript{uncond.} & $51.6 \pm 23.1$ & $\underline{52.2} \pm 16.5$ & $\mathbf{49.2} \pm 20.9$ & $\mathbf{51.0}$ & $\underline{56.1}^\ast \pm 23.3$ & $\mathbf{32.2}$\\
Ours\textsubscript{cond.} & $52.0 \pm 24.5$ & $\mathbf{52.6} \pm 18.6$ & $\underline{44.8} \pm 20.3$ & $\underline{49.8}$ & $\mathbf{57.1} \pm 22.6$ & $\underline{85.4}$\\
\hline
\rowcolor{gray!10} Upper & $60.4 \pm 22.7$ & $53.9 \pm 24.6$ & $56.2 \pm 18.4$ & $56.8$ & $77.3 \pm 13.6$ & $0.0$\\
\hline
\end{tabularx}
\end{table}

\textit{Evaluation Protocol:} To evaluate the quality of the image reconstructions (Cirrus $\to$ Spectralis and Topcon $\to$ Spectralis), we measure the performance of the downstream segmentation models. Specifically, we report the DSC between the networks' predictions and the ground-truth annotations. We assess the segmentation of IRF, SRF, and PED lesions for the RETOUCH dataset, and we quantify the segmentation of GA for the paired in-house dataset. For the paired in-house dataset we also report the Frechet Inception Distance~\cite{HeRa17} (FID) to circumvent the problem of pixel-wise correspondences.

\begin{figure}[ht!]
\centering
\resizebox{\linewidth}{!}{%
\begin{tikzpicture}[
    spy using outlines={
            rectangle, 
            magnification=4, 
            size=0.6cm,          
            connect spies,
            thick
        },
    image_style/.style={inner sep=0, anchor=west}
]

    \node[image_style] (img1) at (0,0) {\includegraphics[width=2cm]{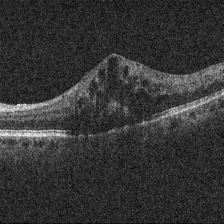}};
    \node[image_style] (img2) [right=0.2cm of img1] {\includegraphics[width=2cm]{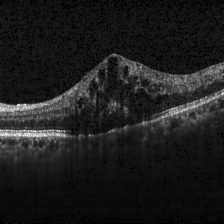}};
    \node[image_style] (img3) [right=0.2cm of img2] {\includegraphics[width=2cm]{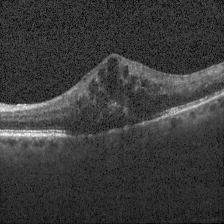}};
    \node[image_style] (img4) [right=0.2cm of img3] {\includegraphics[width=2cm]{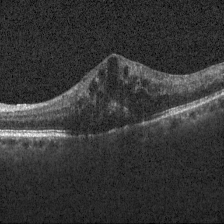}};
    \node[image_style] (img5) [right=0.2cm of img4] {\includegraphics[width=2cm]{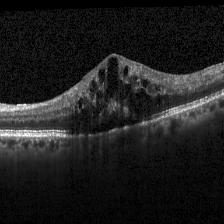}};

    \node[image_style, minimum width=2cm, minimum height=2cm] (img6_empty) [right=0.6cm of img5] {};

    \spy [red, size=1cm] on ($(img1.center) + (-0.6, -0.0)$) in node at ($(img1.south) + (0.45, 0.2)$);
    \spy [red, size=1cm] on ($(img2.center) + (-0.6, -0.0)$) in node at ($(img2.south) + (0.45, 0.2)$);
    \spy [red, size=1cm] on ($(img3.center) + (-0.6, -0.0)$) in node at ($(img3.south) + (0.45, 0.2)$);
    \spy [red, size=1cm] on ($(img4.center) + (-0.6, -0.0)$) in node at ($(img4.south) + (0.45, 0.2)$);
    \spy [red, size=1cm] on ($(img5.center) + (-0.6, -0.0)$) in node at ($(img5.south) + (0.45, 0.2)$);

    \node[image_style] (img6)  [below=0.2cm of img1] {\includegraphics[width=2cm]{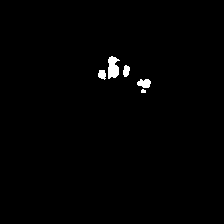}};
    \node[image_style] (img7)  [right=0.2cm of img6]  {\includegraphics[width=2cm]{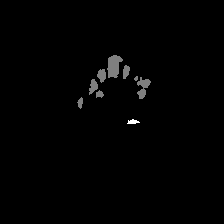}};
    \node[image_style] (img8)  [right=0.2cm of img7]  {\includegraphics[width=2cm]{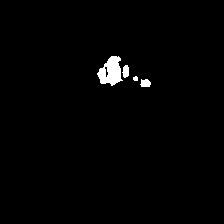}};
    \node[image_style] (img9)  [right=0.2cm of img8]  {\includegraphics[width=2cm]{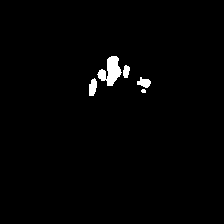}};
    \node[image_style] (img10) [right=0.2cm of img9]  {\includegraphics[width=2cm]{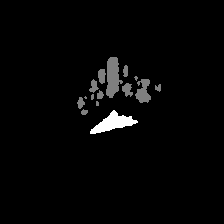}};

    \node[image_style] (img_gt1) [right=0.6cm of img10] {\includegraphics[width=2cm]{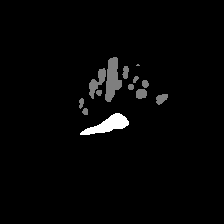}};

    \node at ($(img1.north west)!0.5!(img6.south west) + (-0.7, 0)$) {
    \rotatebox{90}{%
        \shortstack{\large\textbf{Cirrus} \\ \normalsize{RETOUCH}}%
    }
};

    \node[image_style] (img11) [below=0.4cm of img6] {\includegraphics[width=2cm]{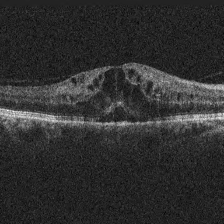}};
    \node[image_style] (img12) [right=0.2cm of img11] {\includegraphics[width=2cm]{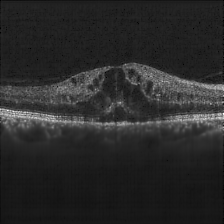}};
    \node[image_style] (img13) [right=0.2cm of img12] {\includegraphics[width=2cm]{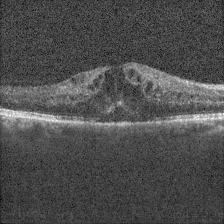}};
    \node[image_style] (img14) [right=0.2cm of img13] {\includegraphics[width=2cm]{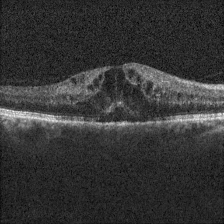}};
    \node[image_style] (img15) [right=0.2cm of img14] {\includegraphics[width=2cm]{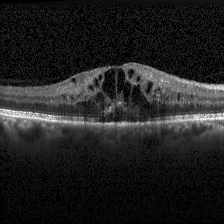}};

    \node[image_style, minimum width=2cm, minimum height=2cm] (img16_empty) [right=0.6cm of img15] {};

    \spy [red, size=1cm] on ($(img11.center) + (-0., 0.3)$) in node at ($(img11.south) + (0.45, 0.2)$);
    \spy [red, size=1cm] on ($(img12.center) + (-0., 0.3)$) in node at ($(img12.south) + (0.45, 0.2)$);
    \spy [red, size=1cm] on ($(img13.center) + (-0., 0.3)$) in node at ($(img13.south) + (0.45, 0.2)$);
    \spy [red, size=1cm] on ($(img14.center) + (-0., 0.3)$) in node at ($(img14.south) + (0.45, 0.2)$);
    \spy [red, size=1cm] on ($(img15.center) + (-0., 0.3)$) in node at ($(img15.south) + (0.45, 0.2)$);

    \node[image_style] (img16)  [below=0.2cm of img11] {\includegraphics[width=2cm]{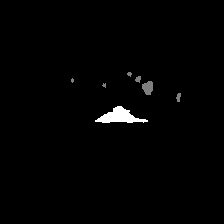}};
    \node[image_style] (img17)  [right=0.2cm of img16]  {\includegraphics[width=2cm]{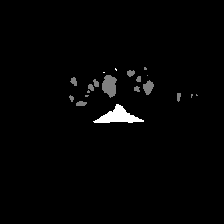}};
    \node[image_style] (img18)  [right=0.2cm of img17]  {\includegraphics[width=2cm]{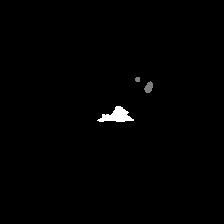}};
    \node[image_style] (img19)  [right=0.2cm of img18]  {\includegraphics[width=2cm]{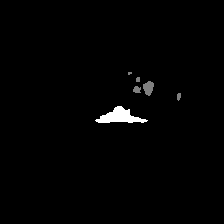}};
    \node[image_style] (img20)  [right=0.2cm of img19]  {\includegraphics[width=2cm]{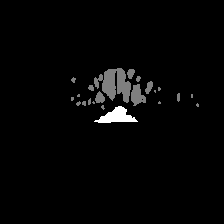}};

    \node[image_style] (img_gt2) [right=0.6cm of img20] {\includegraphics[width=2cm]{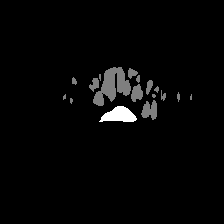}};

    \node[image_style] (img21) [below=0.4cm of img16] {\includegraphics[width=2cm]{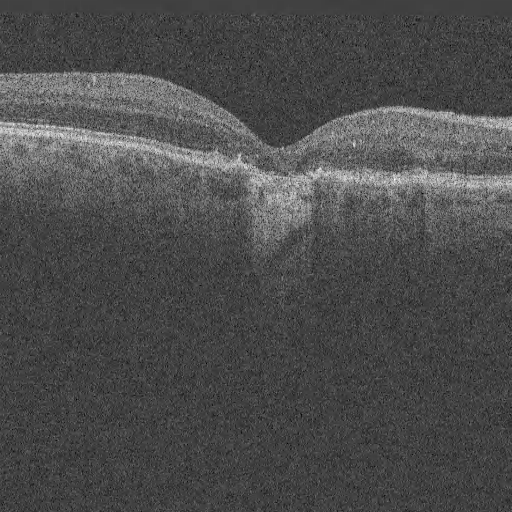}};
    \node[image_style, anchor=center, opacity=0.1] at (img21.center) {\includegraphics[width=2cm]{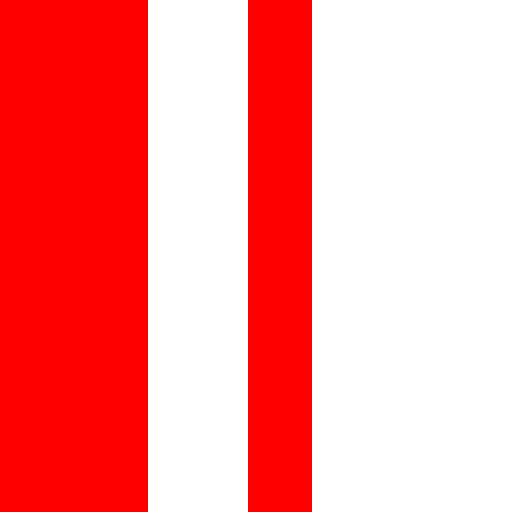}};
    \node[image_style] (img22) [right=0.2cm of img21] {\includegraphics[width=2cm]{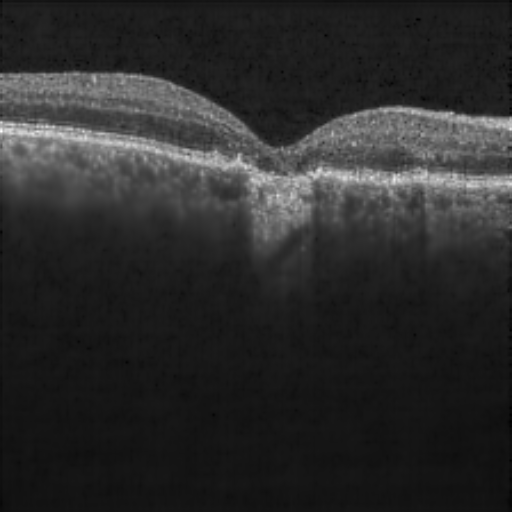}};
    \node[image_style, anchor=center, opacity=0.1] at (img22.center) {\includegraphics[width=2cm]{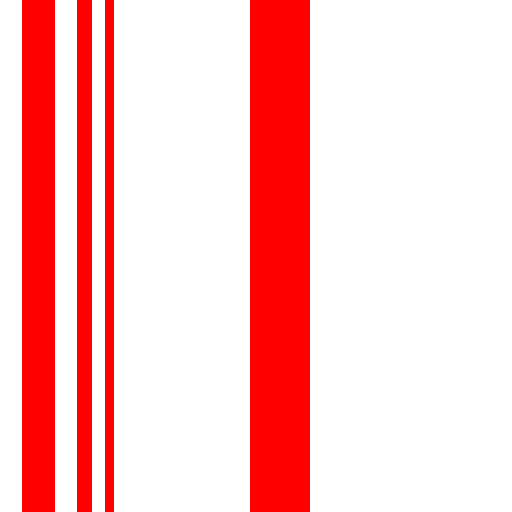}};
    \node[image_style] (img23) [right=0.2cm of img22] {\includegraphics[width=2cm]{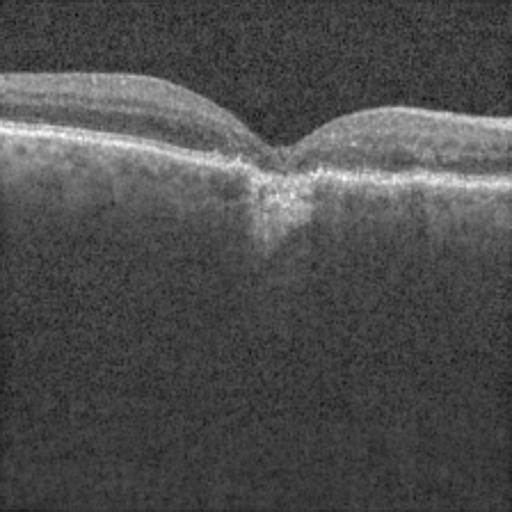}};
    \node[image_style, anchor=center, opacity=0.1] at (img23.center) {\includegraphics[width=2cm]{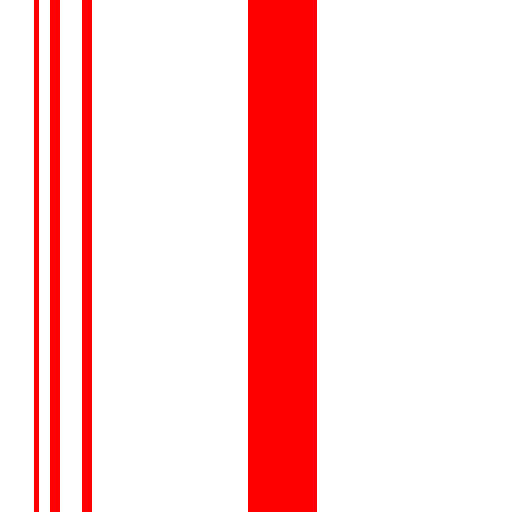}};
    \node[image_style] (img24) [right=0.2cm of img23] {\includegraphics[width=2cm]{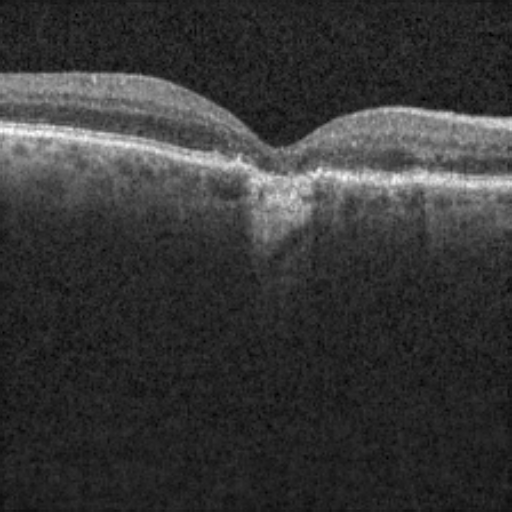}};
    \node[image_style, anchor=center, opacity=0.1] at (img24.center) {\includegraphics[width=2cm]{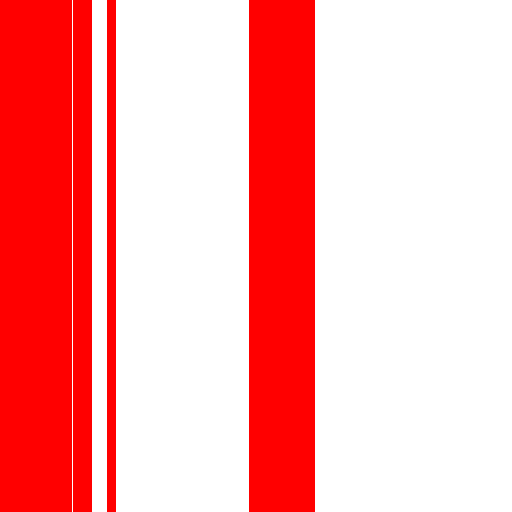}};
    \node[image_style] (img25) [right=0.2cm of img24] {\includegraphics[width=2cm]{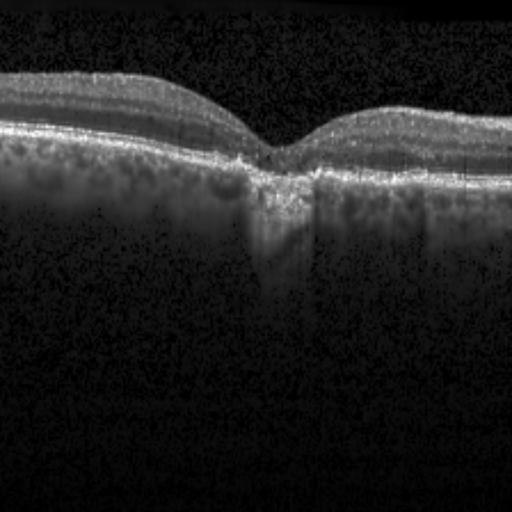}};
    \node[image_style, anchor=center, opacity=0.1] at (img25.center) {\includegraphics[width=2cm]{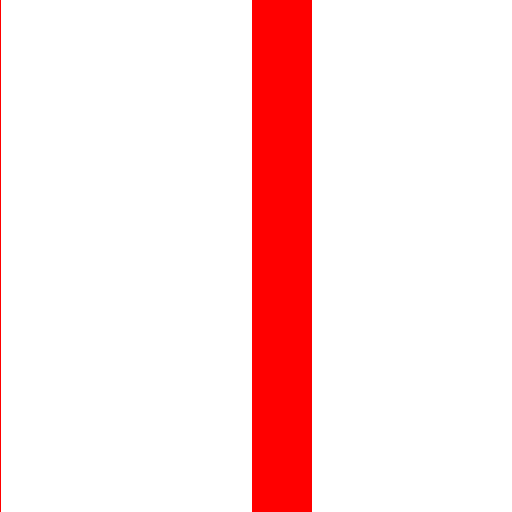}};

    \node[image_style, minimum width=2cm, minimum height=2cm] (img26_empty) [right=0.6cm of img25] {};

    \node[image_style] (img26)  [below=0.2cm of img21] {\includegraphics[width=2cm]{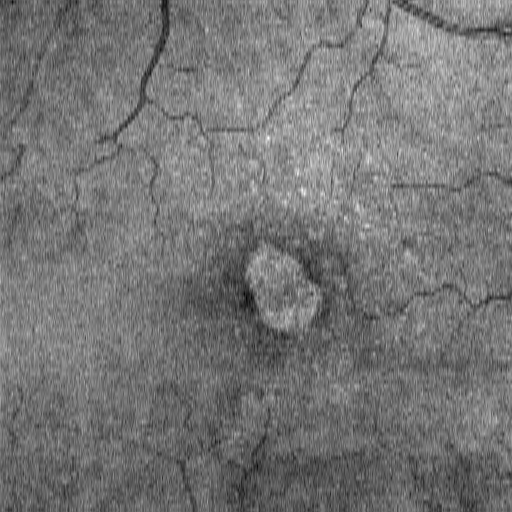}};
    \node[image_style, anchor=center, opacity=0.3] at (img26.center) {\includegraphics[width=2cm]{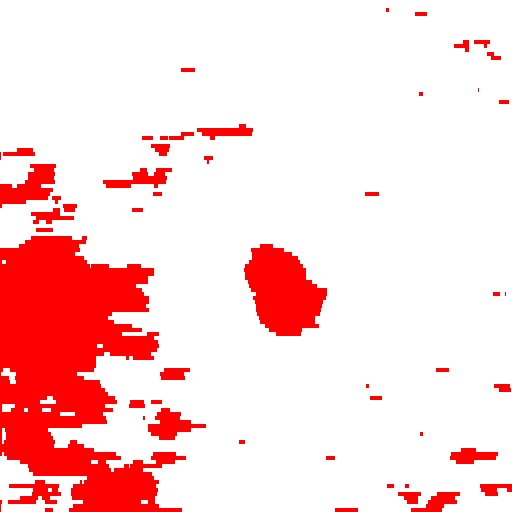}};
    \node[image_style] (img27)  [right=0.2cm of img26]  {\includegraphics[width=2cm]{imgs/GA/example_1/file_28_original_enface.png}};
    \node[image_style, anchor=center, opacity=0.3] at (img27.center) {\includegraphics[width=2cm]{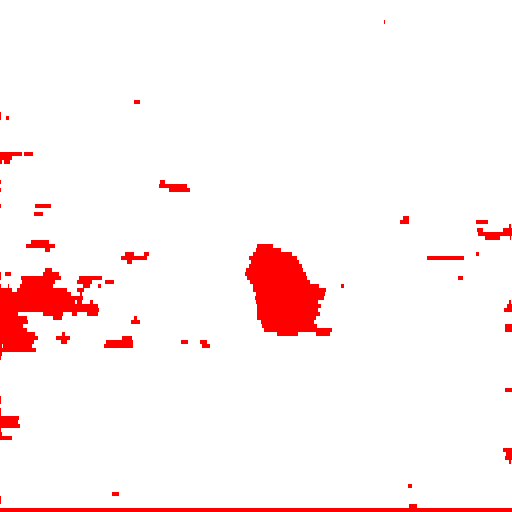}};
    \node[image_style] (img28)  [right=0.2cm of img27]  {\includegraphics[width=2cm]{imgs/GA/example_1/file_28_original_enface.png}};
    \node[image_style, anchor=center, opacity=0.3] at (img28.center) {\includegraphics[width=2cm]{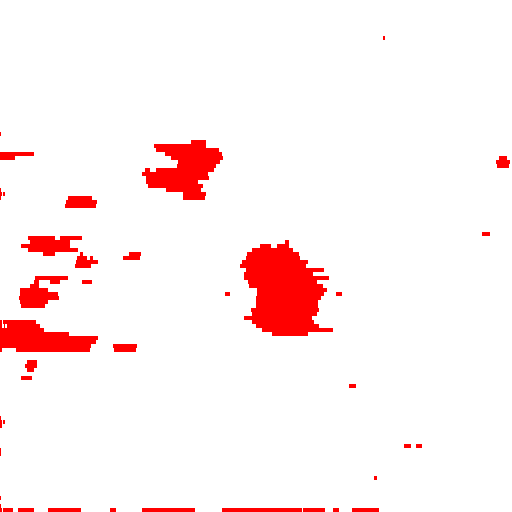}};
    \node[image_style] (img29)  [right=0.2cm of img28]  {\includegraphics[width=2cm]{imgs/GA/example_1/file_28_original_enface.png}};
    \node[image_style, anchor=center, opacity=0.3] at (img29.center) {\includegraphics[width=2cm]{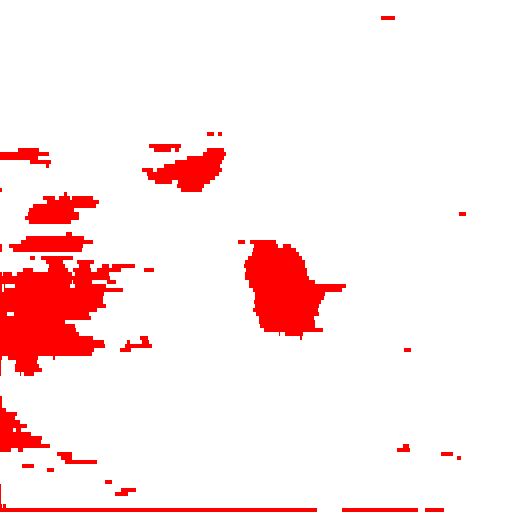}};
    \node[image_style] (img30) [right=0.2cm of img29]  {\includegraphics[width=2cm]{imgs/GA/example_1/file_28_original_enface.png}};
    \node[image_style, anchor=center, opacity=0.3] at (img30.center) {\includegraphics[width=2cm]{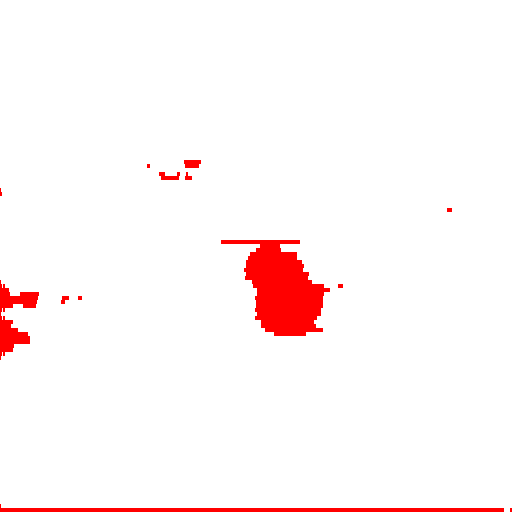}};

    \draw[white, opacity=0.6]
    ($(img26.north west)!0.55!(img26.south west)$) 
    -- 
    ($(img26.north east)!0.55!(img26.south east)$);
    \draw[white, opacity=0.6]
    ($(img27.north west)!0.55!(img27.south west)$) 
    -- 
    ($(img27.north east)!0.55!(img27.south east)$);
    \draw[white, opacity=0.6]
    ($(img28.north west)!0.55!(img28.south west)$) 
    -- 
    ($(img28.north east)!0.55!(img28.south east)$);
    \draw[white, opacity=0.6]
    ($(img29.north west)!0.55!(img29.south west)$) 
    -- 
    ($(img29.north east)!0.55!(img29.south east)$);
    \draw[white, opacity=0.6]
    ($(img30.north west)!0.55!(img30.south west)$) 
    -- 
    ($(img30.north east)!0.55!(img30.south east)$);

    \node at ($(img11.north west)!0.5!(img26.south west) + (-1., 0)$) {
    \rotatebox{90}{%
        \shortstack{\large\textbf{Topcon}}%
    }
    };
    \node at ($(img11.north west)!0.5!(img16.south west) + (-0.5, 0)$) 
        {\rotatebox{90}{\normalsize{RETOUCH}}};
    \node at ($(img21.north west)!0.5!(img26.south west) + (-0.5, 0)$) 
    {\rotatebox{90}{\normalsize{In-house}}};

    \node[image_style] (img_gt3) [right=0.6cm of img30] {\includegraphics[width=2cm]{imgs/GA/example_1/file_28_original_enface.png}};
    \node[image_style, anchor=center, opacity=0.3] at (img_gt3.center) {\includegraphics[width=2cm]{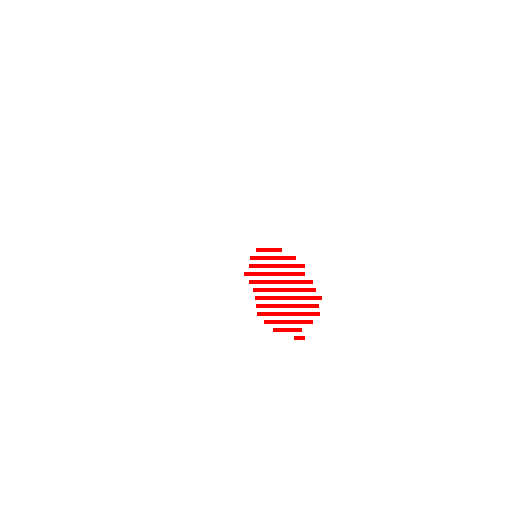}};

    \draw[thick, gray!75]
    ($(img6.south west)!0.5!(img11.north west) + (-0.5, 0)$)
    --
    ($(img_gt1.south east)!0.5!(img16_empty.north east) + (0.3, 0)$);

    \draw[thick, gray!75]
    ($(img16.south west)!0.5!(img21.north west) + (-0.5, 0)$)
    --
    ($(img_gt2.south east)!0.5!(img26_empty.north east) + (0.3, 0)$);

    \draw[thick, gray!75] 
    ($(img5.north east)!0.5!(img6_empty.north west) + (0, 0.2)$) 
    -- 
    ($(img30.south east)!0.5!(img_gt3.south west) + (0, -0.4)$);

    \node[below=0.1cm of img26] {Baseline};
    \node[below=0.1cm of img27] {Flow Matching};
    \node[below=0.1cm of img28] {DDA};
    \node[below=0.1cm of img29] {CPDM};
    \node[below=0.1cm of img30] {Ours};
    \node[below=0.1cm of img_gt3] {GT};

\end{tikzpicture}
}
\caption{Comparison to the state-of-the-art on downstream fluid and GA segmentation on the \textbf{Cirrus} (first two rows) and \textbf{Topcon} (remaining rows) data. For the GA segmentation, a white horizontal line denotes the position of the corresponding bscan (In-house, top row). Note that only every second bscan is annotated in the GA GT.}
\label{fig:qualitative_comparison}
\end{figure}

\section{Results and Discussion}
\textit{Image reconstruction:} 
Table~\ref{tab:merged_performance} summarizes the downstream segmentation performance for the Cirrus $\to$ Spectralis and Topcon $\to$ Spectralis reconstruction tasks, respectively. On the Cirrus data, our unconditional method achieves a state-of-the-art mean DSC of 58.6, significantly outperforming the baseline methods, and approaching the supervised upper bound.

The Topcon $\to$ Spectralis shift appears to be more challenging, indicated by systematically lower scores. Despite this difficulty, our approach maintains the highest mean DSC (51.0) on the RETOUCH dataset. While TENT slightly outperforms on IRF lesions, it fails to generalize to other structures like PED, where our method remains stable. Fig.~\ref{fig:qualitative_comparison} provides a qualitative comparison of our reconstructions against baseline generative methods. For both Cirrus and Topcon data, our approach yields high-quality reconstructions that preserve the original anatomy while visibly reducing noise and maintaining high contrast. The improved reconstruction translates to the downstream task, resulting in fluid segmentations that align well with the ground-truth annotations. Similarly, for the GA task (Fig.~\ref{fig:qualitative_comparison}, bottom), our approach notably reduces segmentation errors compared to all baselines, demonstrating a significant improvement over its closest competitor DDA.

\begin{figure}[tb]
    \includegraphics[width=\textwidth]{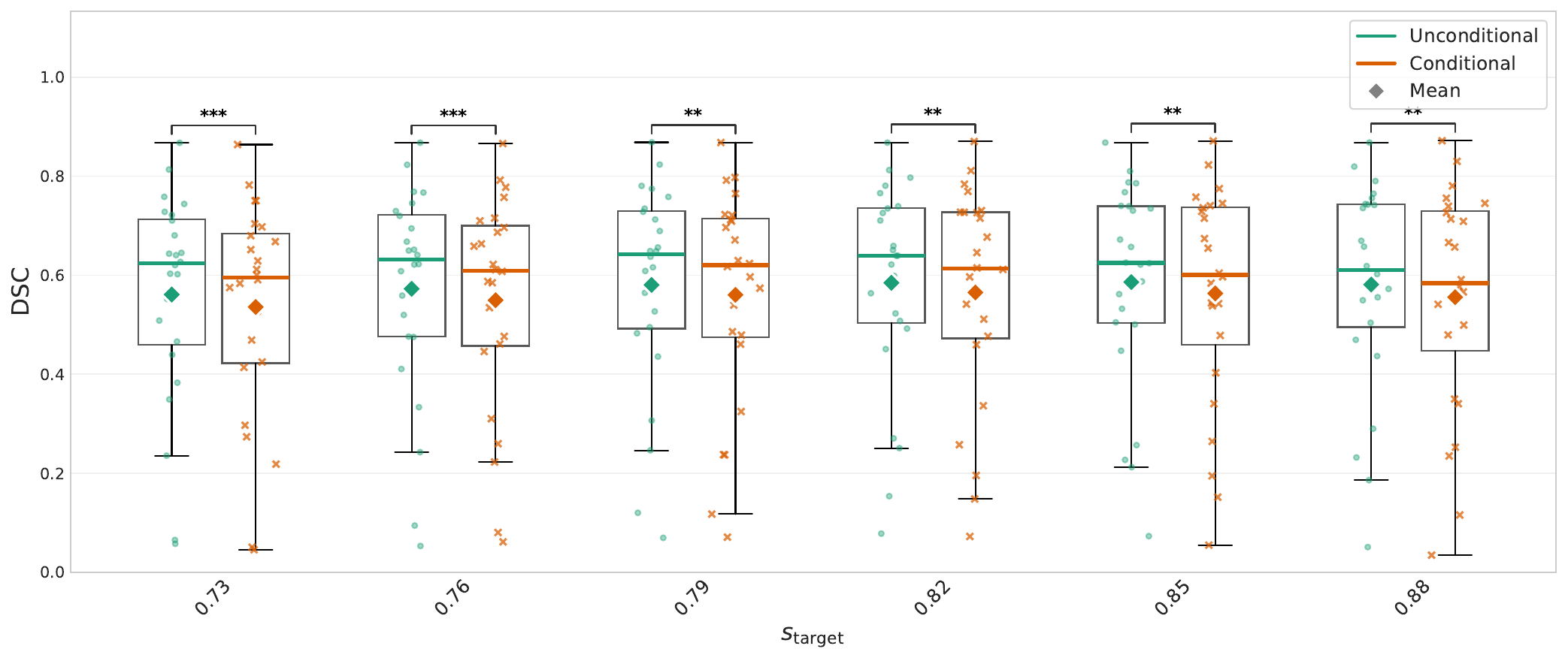}
    \caption{Comparison of downstream lesion segmentation performance (DSC) for different reference time points $s_{\text{target}}$. For each target, the results of the conditional and unconditional networks are shown. *denotes statistical significance.}
    \label{fig:boxplots_conditional_unconditional_cirrus}
\end{figure}

\textit{Ablation Study:} 
To validate our architectural design, we evaluate the impact of the histogram matching module and the time-conditioning parameterization. 
First, integrating the histogram matching module provides a significant performance gain over plain Flow Matching (FM), improving the mean DSC from 50.8 to 58.6 and from 49.4 to 56.1 on the Cirrus and the in-house data, respectively. 
Second, comparing the noise-conditional (Ours\textsubscript{cond.}) and unconditional (Ours\textsubscript{uncond.}) networks reveals a trade-off in Table~\ref{tab:merged_performance}. Dropping the time-conditioning forces the network to infer the noise state implicitly, yielding better overall fluid segmentation. As illustrated in Fig.~\ref{fig:boxplots_conditional_unconditional_cirrus}, this unconditional approach consistently outperforms the conditional version and exhibits low sensitivity to the starting point $s_{\text{target}}$. In fact, oracle selection (choosing the absolute best hyperparameter per individual patient) improves overall results by only 1 percentage point. Additionally, the unconditioned model produces superior perceptual quality, reducing the FID from 85.4 to 32.2. However, the conditional model offers a non-significant benefit on the specific GA task (DSC 57.1 vs. 56.1).

\section{Conclusion}
In this study, we introduced a novel, Flow Matching-based framework TTA-Flow for high-quality OCT image reconstruction.
We demonstrated that histogram matching is a simple yet highly effective preprocessing technique for aligning the intensity distribution of test data with the network's expectations.
Because our reconstruction process integrates the probability-flow ODE of a generative model, the required reference histograms can be extracted at virtually no additional computational cost from synthetic trajectories.
Furthermore, our results indicate that removing time-conditioning from the flow-matching network enhances denoising performance, leading to superior fluid segmentation and visual fidelity compared to standard conditional setups.
While our setup without the usage of data-fidelity terms can theoretically hallucinate structures, we did not observe this in practice and data-fidelity terms are very sensitive to the hyperparameters. Notably, our method outperforms multiple baselines that explicitly use data-fidelity terms, including CPDM~\cite{li2024score}, which utilizes a fidelity term strictly designed for OCT. 
Currently, the optimal reference time point for histogram matching is set globally as a fixed hyperparameter.
However, our future work will explore dynamically estimating the noise level directly from the input image as well as speeding up the adaptation process.
This advancement will enable the framework to adaptively choose the ideal histogram-matching target for each volume, fully automating the pipeline and optimizing reconstruction fidelity.

\paragraph{Acknowledgements}: Funded by the European Union (I-SCREEN, grant no. 101130093), EIC-2023-PATHFINDEROPEN-01 . Views and opinions expressed are however those of the author(s) only and do not necessarily reflect those of the European Union or European Innovation Council and SMEs Executive Agency (EISMEA). Neither the European Union nor the granting authority can be held responsible for them.

\paragraph{Disclosure of Interests.}  The authors have no competing interests to declare.

%
%
%
\bibliographystyle{splncs04}
\bibliography{paper-1314}

\end{document}